\documentclass{article}
\usepackage[final]{lmca2020}

\usepackage[utf8]{inputenc} 
\usepackage[T1]{fontenc}    
\usepackage{hyperref}       
\usepackage{url}            
\usepackage{booktabs}       
\usepackage{amsfonts}       
\usepackage{nicefrac}       
\usepackage{microtype}      
\usepackage{xcolor}
\usepackage{soul}  
\usepackage{listings}  
\usepackage{graphicx}  
\usepackage{courier}
\usepackage{todonotes}

\graphicspath{{./figures/}}

\setcitestyle{numbers,square}

\definecolor{codegreen}{rgb}{0,0.6,0}
\definecolor{codegray}{rgb}{0.5,0.5,0.5}
\definecolor{codepurple}{rgb}{0.58,0,0.82}
\definecolor{backcolour}{rgb}{0.95,0.95,0.92}

\lstdefinestyle{mystyle}{
  backgroundcolor=\color{backcolour},
  commentstyle=\color{codegreen},
  keywordstyle=\color{magenta},
  numberstyle=\tiny\color{codegray},
  stringstyle=\color{codepurple},
  basicstyle=\ttfamily\scriptsize,
  breakatwhitespace=false,
  breaklines=true,
  captionpos=b,
  keepspaces=true,
  numbers=left,
  numbersep=5pt,
  showspaces=false,
  showstringspaces=false,
  showtabs=false,
  tabsize=2
}

\newcommand{\code}[1]{{\footnotesize\texttt{#1}}}
\title{Ecole: A Gym-like Library for Machine Learning in Combinatorial Optimization Solvers}

\author{%
    Antoine Prouvost \\
    Mila, Polytechnique Montréal \\
   \And
    Justin Dumouchelle \\
    Polytechnique Montréal \\
   \And
    Lara Scavuzzo \\
    Technische Universiteit Delft \\
   \And
    Maxime Gasse \\
    Mila, Polytechnique Montréal \\
   \And
    Didier Chételat \\
    Polytechnique Montréal \\
   \And
   Andrea Lodi \\
    Mila, Polytechnique Montréal \\
}

\date{September 2020}

\begin{document}

\maketitle

\begin{abstract}
We present Ecole, a new library to simplify machine learning research for combinatorial optimization. Ecole exposes several key decision tasks arising in general-purpose combinatorial optimization solvers as control problems over Markov decision processes. Its interface mimics the popular OpenAI Gym library and is both extensible and intuitive to use. We aim at making this library a standardized platform that will lower the bar of entry and accelerate innovation in the field. Documentation and code can be found at \url{https://www.ecole.ai}. 
\end{abstract}

\section{Introduction}

In many industrial applications, such as day-to-day lot sizing and production planning~\cite{book/PochetW06}, it is common to repeatedly solve similar NP-hard combinatorial optimization (CO) problems. In practice those are typically fed into an off-the-shelf, general-purpose mathematical solver, which processes each new problem independently and retains no memory of the past. Yet, it is very likely that there exist strong statistical similarities between each of those sequentially solved problems, which could potentially be exploited to solve future problems more efficiently. This observation has motivated two growing lines of machine learning research: 1) pure machine learning (ML) approaches, where CO solvers are entirely replaced by an ML model trained to produce (near)-optimal solutions~\cite{confs/iclr/BelloPLNB17,conf/nips/DaiKZDL17,conf/iclr/KoolVHW19}; and 2) joint approaches, where hand-designed decision criteria within classical CO solvers are replaced by machine learning models trained to optimize a particular metric of the solver \cite{bengio2020machine}. The latter approach is particularly attractive as it allows for exact solving, or at least for mathematical guarantees on the optimality gap (lower and upper bounds), which are often of high value in practice.

Leading general-purpose CO solvers such as Gurobi \cite{software/gurobi}, IBM CPLEX \cite{software/cplex}, FICO Xpress~\cite{journal/annalsor/Robert07} or SCIP \cite{software/scip7} are all based on the branch-and-cut algorithm \cite{study/AchterbergW13,book/Grotschel04}. This algorithm iteratively divides the feasible space and prunes away sections of that space that cannot contain the optimum using bounds derived from linear programming (LP) relaxations strengthened with cuts \cite{article/hao/Mitchell02}. At many points during the algorithm, decisions must be taken that greatly impact the solving performance, traditionally by following a series of hand-crafted rules designed by operations research (OR) experts. Thus, a natural direction to improve the performance of these solvers is to replace the hand-crafted decision rules by machine learning models, trained on representative problems. This promising line of research has already shown improvement on several of these decision tasks, including variable selection~\cite{journals/ejor/DiLibertoKLM2016,conf/aaai/KhalilBSND16,journal/informs-joc/AlvarezLW17,conf/icml/BalcanDSV18,arxiv/HansknechtJS2018,conf/neurips/GasseCFCL19,arxiv/zarpellon2020parameterizing},
node selection~\cite{conf/nips/HeDE14, arxiv/SongLZYO18,conf/cpaior/sabharwal2012guiding},
cut generation~\cite{conf/icml/TangAF20},
column generation (a.k.a. pricing)~\cite{gerad/MorabitDL20}, 
primal heuristic selection~\cite{conf/ijcai/KhalilDNAS17,conf/gor/HendelMW18}, or formulation selection~\cite{conf/cpaior/BonamiLZ18}.

All these works have in common that learning can be formulated as a control problem over a Markov decision process (MDP), where a branch-and-cut solver constitutes the environment. Such a formulation opens the door to reinforcement learning (RL) algorithms, which have been successful in solving extremely complex tasks in other fields~\cite{journal/science/alphazero,journals/nature/alphastar}. These data-driven policies may hopefully improve upon the expert heuristics currently implemented in commercial solvers, and by doing so highlight new research directions for the combinatorial optimization community.

\section{Motivation}
\label{sec:challenges}

Although the idea of using ML for decision-making within CO solvers is receiving increasing attention, research in this area also suffers from several unfortunate technical obstacles, which hinders scientific progress and innovation.

First, reproducibility is currently a major issue. The variety of solvers, problem benchmarks, hand-crafted features, and evaluation metrics used in existing studies impedes reproducibility and comparison to previous works. Those same issues have driven the ML and RL communities to adopt standardized evaluation benchmarks, such as ImageNet \cite{confs/cvpr/DengDSLLF09} or the Arcade Learning Environment \cite{journals/jair/BellemareNVB13}. We believe that adopting standard feature sets, problem benchmarks and evaluation metrics for several identified key problems (e.g., branching, node selection, cutting plane generation) will be highly beneficial to this research area as well.

Second, there is a high bar of entry to the field. Modern solvers are complex pieces of software whose implementation always deviates from the vanilla textbook algorithm, and which were not specifically designed for direct customization through machine learning.
Implementing a new research idea often requires months of digging in the technical intricacies of low-level C solver code, even for OR experts, and requires ML experts joining the field to be very familiar with the inner working of a CO solver. On the other hand, abstracting away a proper MDP formulation using a solver API is no trivial task either for OR experts, and requires a clear understanding of statistical learning concepts and their significance. We believe that exposing several decision tasks of interest through a unified ML-compatible API will help attract interest from both the traditional ML and OR communities.

Finally, at this time the field hardly benefits from the latest advances in both ML and OR. ML experts typically employ very simplified CO solvers or no solver at all~\cite{conf/icml/TangAF20,conf/iclr/KoolVHW19}, raising criticism among the OR community, while OR experts typically employ basic ML models and algorithms~\cite{journal/informs-joc/AlvarezLW17,conf/cpaior/BonamiLZ18,conf/gor/HendelMW18}, thereby missing potential improvements. We believe that a plug-and-play API between a state-of-the-art CO solver and ML algorithms, in the form of a Gym-compatible interface, will allow for closing this gap and let the field benefit from the latest advances from both sides.

\section{Proposed solution}

To address these practical challenges, we propose a novel open-source library that could serve as a universal platform for research and development in the ML within CO. This new platform, the \emph{Extensible Combinatorial Optimization Learning Environments} (Ecole) library, is designed as an interface between a CO solver and ML algorithms. It provides a collection of key decision tasks, such as variable selection or cut selection, as partially-observable (PO)-MDP environments in a way that closely mimics OpenAI Gym \cite{arxiv/BrockmanCPSSTZ16}, a widely popular library among the RL community.

\subsection{Design}

\begin{figure}[t]
\centering
\begin{lstlisting}
import ecole

# set up an MDP environment
env = ecole.environment.Branching(
    # use the features from Gasse et al., 2019
    observation_function=ecole.observation.NodeBipartite(),
    # minimize the B&B tree size
    reward_function=-ecole.reward.NNodes())

# set up an instance generator
instances = ecole.instance.CombinatorialAuctionGenerator(n_items=100, n_bids=100)

# generate ten MDP episodes
for _ in range(10):
    # new instances are generated on-the-fly
    instance = next(instances)
    # save instance to disk if desired
    instance.write_problem(f"path/to/problem_{i}.lp")
    # start a new episode
    obs, action_set, reward, done = env.reset(instance)
    # unroll the control loop until the instance is solved
    while not done:
        action = ...  # decide on the next action here
        obs, action_set, reward, done, info = env.step(action)
\end{lstlisting}
\caption{Example code snippet, using Ecole for branching on combinatorial auction problems.}
\label{fig:example_code}
\end{figure}

The design of the library was guided to achieve the following objectives.

\paragraph{Modularity} %
An environment in Ecole is defined by a composition of a task, an observation function and a reward function. For example, in Figure~\ref{fig:example_code}, a branching environment is defined with a node bipartite graph observation and the negative number of new nodes created as a reward. Users can define their own observation or reward function to fulfill their specific needs, or even define new environments and simply reuse existing observation and reward functions. These new modules can be defined either directly in C++ for speed, or in Python for flexibility.

\paragraph{Scalability} %
Ecole was designed to add as little overhead as possible on top of the solver. In addition, care was taken to ensure that the library is thread-safe, and in particular Ecole was designed to be free from the Python Global Interpreter Lock (GIL). This allows for straightforward parallelism in Python with multi-threading, which simplifies data collection and policy evaluation in RL algorithms.

\paragraph{Speed} %
The Ecole core is written in C++, interacts directly with the low-level solver API and provides a thin Python API returning Numpy arrays~\cite{numpy11} to interface directly with ML libraries. The initial release of Ecole supports the state-of-the-art open-source solver SCIP~\cite{software/scip7} as a backend, due to its open code that gives complete access to the solver, and its widespread usage in the literature \cite{conf/nips/HeDE14, arxiv/HansknechtJS2018, conf/neurips/GasseCFCL19, arxiv/DingZSLWXS19, arxiv/YilmazYS20, conf/neurips/GuptaGKKLB20}. We hope in future versions to expand the library to other commercial solvers as well, such as Gurobi~\cite{software/gurobi}, CPLEX~\cite{software/cplex} or Xpress \cite{journal/annalsor/Robert07} if the developers of these solvers are interested.

\paragraph{Flexibility} %
Ecole can read instance files in any format understood by SCIP and can therefore be used with existing benchmark collections such as MIPLIB~\cite{miplib2017}.
In addition, the library also provides out-of-the box instance generators for classical CO problems, which can be used to quickly test ideas or to offer standard benchmarks. The generated instances can be saved to disk or passed directly to Ecole environments from memory, as illustrated in Figure~\ref{fig:example_code}. Four instance generators are currently implemented (combinatorial auctions, maximum independent set, capacitated facility location and set covering problems) with default parameters chosen to yield solving times on the order of a minute.

\paragraph{Openness} To encourage widespread usage of the library, we chose to distribute Ecole under an open-source BSD-3 license \cite{misc/bsd3}. In addition, care was taken that the library could be installed with the popular \code{conda} package manager \cite{software/conda}, which is widely used in the ML community.

\subsection{Supported features}

The library currently supports two control tasks. The first is hyperparameter tuning (\code{ecole.environment.Configuring}), the task of selecting the best solver hyperparameters before solving. The second is variable selection (\code{ecole.environment.Branching}), the task of deciding sequentially on the next variable to branch on during the construction of the branch-and-bound tree. The library also includes an empty ``baseline'' environment that can be used to benchmark against the solver in its default settings. We are actively working on expanding the library with several other environments that correspond to key research questions in the field, such as node selection and cut selection.

In addition, the library currently supports two observation functions for the state of the solving process. The first is the finite-dimensional variable-aggregated representation from \citet{conf/aaai/KhalilBSND16}, and the second is the bipartite graphical representation from~\citet{conf/neurips/GasseCFCL19}. Finally, the library currently supports two standard reward functions, namely the number of branch-and-bound nodes, and the number of LP iterations added since the last decision. Several standard metrics will be added, such as the primal integral, the dual integral, the primal-dual integral, and the solving time.

\subsection{Example use case}

\begin{figure}[t]
    \centering
    \includegraphics[scale=0.5]{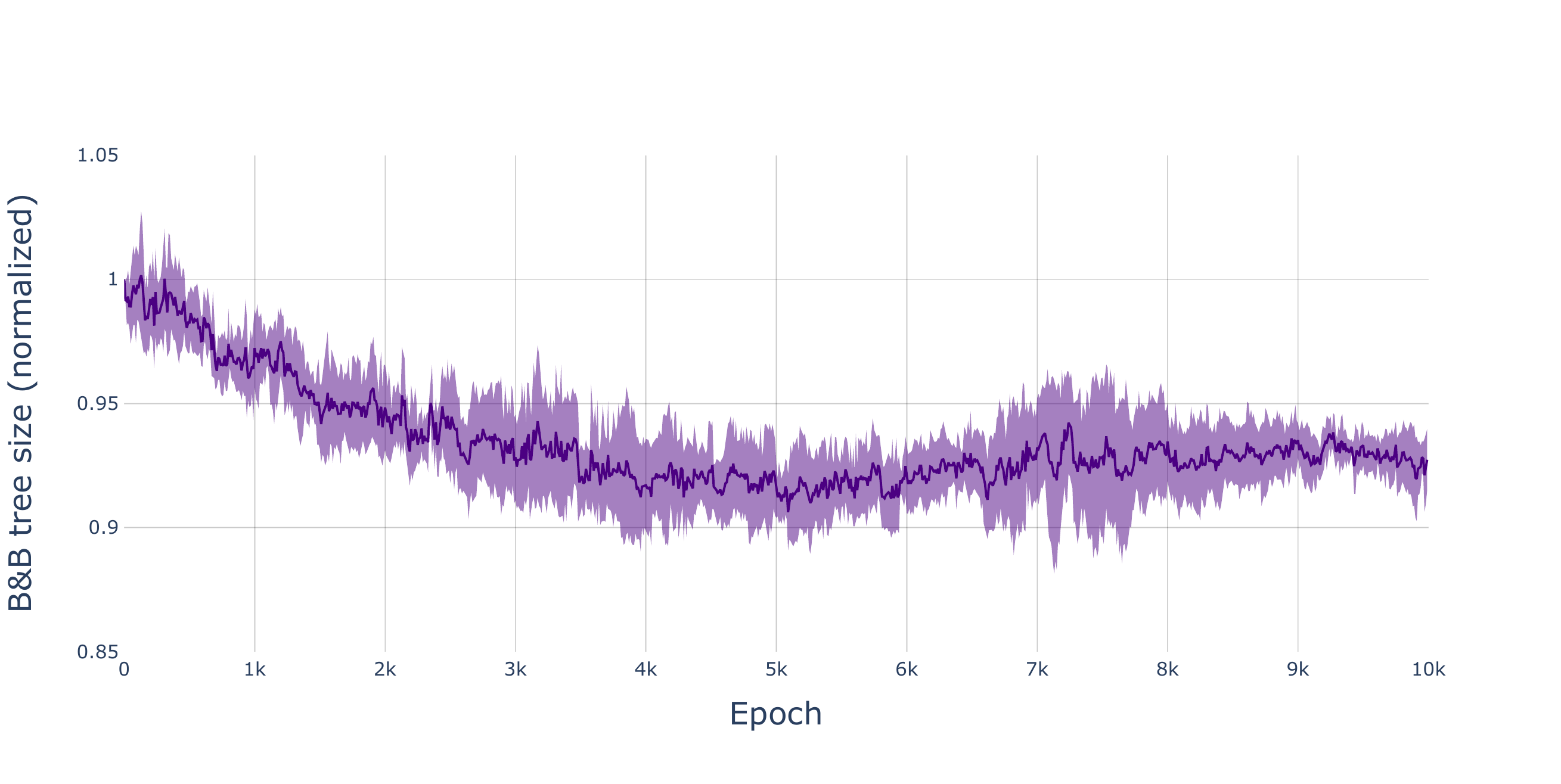}
    \caption{Training curve of a branching environment on randomly generated Combinatorial Auction instances. We report the normalized performance on validation instances, the lower the better.}
    \label{fig:rl_curve}
\end{figure}

We now demonstrate a typical use case, namely reinforcement learning for variable selection in branch and bound. We used the negative number of nodes created between two decisions as reward function, and the policy has the same architecture as the GNN model of \citet{conf/neurips/GasseCFCL19}. We pretrained the weights by imitation learning of strong branching, as in the cited paper, and further trained the policy using REINFORCE \cite{journals/machinelearning/Williams92} on samples of transitions encountered during rollouts. Figure \ref{fig:rl_curve} shows a 5-10\% decrease in the branch-and-bound tree size after 10k episodes. Most interestingly for this article, the environment was defined with only a few lines of code using Ecole, namely a combination of \code{ecole.environment.Branching}, \code{ecole.observation.NodeBipartite} and \code{ecole.reward.NNodes}, and the code was parallelized on 8 threads using the native \code{threading} Python library. Equivalent code in PySCIPOpt \cite{software/pyscipopt}, the SCIP Python API, would have been substantially more complex to write.

\section{Related work}

Other open-source libraries have been recently proposed to simplify research at the intersection of machine learning and combinatorial optimization. MIPLearn \cite{software/miplearn} is a customizable library for machine-learning-based solver configuration currently supporting Gurobi and CPLEX. It offers similar functionalities to the configuration environment in Ecole, which is a (potentially contextual) bandit problem, and can be framed as a borderline case of our MDP framework. In addition, ORGym \cite{software/orgym} and OpenGraphGym \cite{software/opengraphgym} are Gym-like libraries for learning heuristics for a collection of combinatorial optimization problems that are formulated as sequential decision making problems. Thus, in those libraries there is an explicit MDP formulation like in Ecole, although in those the goal is to replace CO solvers entirely, while Ecole aims at improving existing CO solvers. As such, none of these libraries has the ambitious objective of Ecole, which is to serve as a standardized platform for ML within CO solvers.

\section{Conclusions}

In this paper, we proposed a new open-source library that offers Gym-like Markov decision process interfaces to key decision tasks in combinatorial optimization solvers. This library was designed to be fast, modular, scalable, and flexible, with easy installation. Such a library is intended to improve reproducibility, lower the bar of entry and simplify integration of recent advances from both fields, in this growing area at the intersection of machine learning and combinatorial optimization.

\section*{Acknowledgements}

This work was supported by the Canada Excellence Research Chair (CERC) in Data Science for Real-Time Decision Making and IVADO.

\bibliography{biblio}

\end{document}